\documentclass[letterpaper, 10 pt, conference]{ieeeconf}  % Comment this line out if you need a4paper

\IEEEoverridecommandlockouts                              % This command is only needed if 
\makeatletter
\let\NAT@parse\undefined
\makeatother

\usepackage{cite}
\usepackage{amsmath,amssymb,amsfonts}
\usepackage{graphicx}
\usepackage{textcomp}
\usepackage{multirow}
\usepackage{colortbl}
\usepackage{caption}
\usepackage{subcaption}
\usepackage{booktabs}
\usepackage{float}
\usepackage{newfloat}
\usepackage[skins]{tcolorbox}
\usepackage{hyperref}
\usepackage{tikz}

\newcommand\copyrighttext{%
  \footnotesize \copyright~2025 IEEE. Personal use of this material is permitted.  Permission from IEEE must be obtained for all other uses, in any current or future media, including reprinting/republishing this material for advertising or promotional purposes, creating new collective works, for resale or redistribution to servers or lists, or reuse of any copyrighted component of this work in other works.}
\newcommand\copyrightnotice{%
\begin{tikzpicture}[remember picture,overlay]
\node[anchor=south,yshift=10pt] at (current page.south) {\parbox{\dimexpr\textwidth-\fboxsep-\fboxrule\relax}{\copyrighttext}};
\end{tikzpicture}%
}

\newfloat{promptfloat}{!t}{pop}
\newtcolorbox[auto counter]{prompt}[2]{
    coltitle=black,
    label={prompt:#1},
    colback=gray!10,
    colframe=black!0!black,
    fonttitle=\bfseries,
    title=Prompt \thetcbcounter: #2,
    enhanced,
    fonttitle=\scriptsize, fontupper=\scriptsize, fontlower=\scriptsize, left=1mm, 
    right=1mm, 
    top=1mm, 
    bottom=1mm, 
    middle=1mm,
    arc=0pt,
    boxrule=0pt,
    borderline={1pt}{0pt}{dashed}, % Border style
    minipage boxed title*=-1.95em,
    attach boxed title to bottom center={yshift=2pt, yshift=0pt},
    boxed title style={enhanced, colback=white!55!white,
    boxrule=0pt, frame hidden}
}

\title{\LARGE \bf
Shaken, Not Stirred: A Novel Dataset for Visual Understanding of Glasses in Human-Robot Bartending Tasks
}

\author{Lukáš Gajdošech$^{1\dagger}$, Hassan Ali$^{2\dagger}$, Jan-Gerrit Habekost$^{2\dagger}$, Martin Madaras$^{1}$, Matthias Kerzel$^{2}$, \\and Stefan Wermter$^{2}$% <-this % stops a space
\thanks{$^\dagger$These authors contributed equally to this work.}%
\thanks{*Work supported by Horizon Europe project TERAIS (Grant number 101079338) and the DFG Crossmodal Learning (TRR-169) project.}% <-this % stops a space
\thanks{$^{1}$Lukáš Gajdošech and Martin Madaras are with Department of Applied Informatics, Faculty of Mathematics, Physics and Informatics, Comenius University, Bratislava, Slovakia
        {\tt\small \{lukas.gajdosech,martin.madaras\}@fmph.uniba.sk}}%
\thanks{$^{2}$Hassan Ali, Jan-Gerrit Habekost, Matthias Kerzel and Stefan Wermter are with the Knowledge Technology Group, Department of Informatics, University of Hamburg,
        Hamburg, Germany
        {\tt\small \{hassan.ali,jan-gerrit.habekost,matthias.kerz\newline
        el, stefan.wermter\}} {\tt\small @uni-hamburg.de}}%
}

\begin{document}

\makeatletter
\g@addto@macro\@maketitle{
  \captionsetup{type=figure}\setcounter{figure}{0}
  \def\mycolspace{1.2mm}
  \centering
    \includegraphics[trim=0.0 0.0 0.0 0, clip, width=1.63\columnwidth]{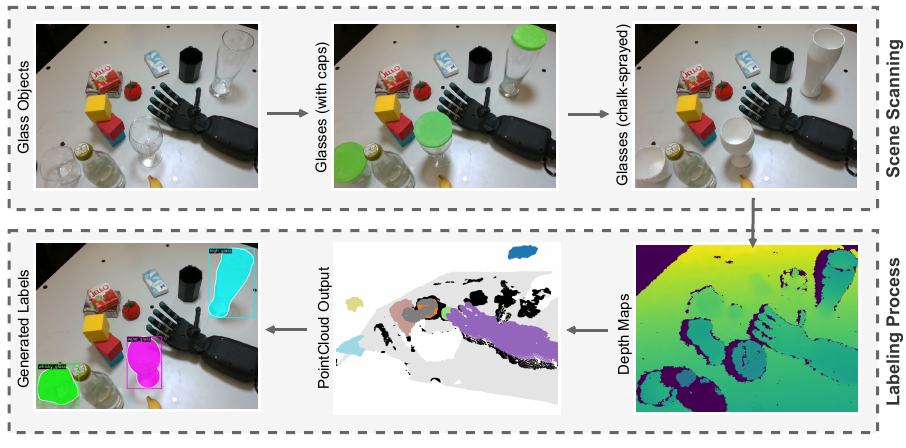}
    %\vspace{-1ex}
        \captionof{figure}{An overview of our proposed auto-labeling pipeline: Each scene is scanned in three stages% (glasses, glasses with caps, and chalk-sprayed glasses)
        . Depth maps and point clouds are generated using RGB-D sensors, followed by %an automatic labeling that combines 
        height verification, color matching, and object detection. 
        \label{fig:pipe}}
        %\vspace{-3ex}
}

\maketitle
\thispagestyle{empty}
\pagestyle{empty}

%\begin{figure*}[t!]
%    \centering
%    \includegraphics[width=0.95\textwidth]{imgs/fusion.png}   
%    \caption{Fusion of all RGB-D Camera Views}
%    \label{fig:enter-label}
%\end{figure*}

%%%%%%%%%%%%%%%%%%%%%%%%%%%%%%%%%%%%%%%%%%%%%%%%%%%%%%%%%%%%%%%%%%%%%%%%%%%%%%%%

\begin{abstract}
Datasets for object detection often do not account for enough variety of glasses, due to their transparent and reflective properties. Specifically, open-vocabulary object detectors, widely used in embodied robotic agents, fail to distinguish subclasses of glasses. This scientific gap poses an issue for robotic applications that suffer from accumulating errors between detection, planning, and action execution. This paper introduces a novel method for acquiring real-world data from RGB-D sensors that minimizes human effort. We propose an auto-labeling pipeline that generates labels for all the acquired frames based on the depth measurements. We provide a novel real-world glass object dataset\footnote[3]{\footnotesize\href{https://knowledgetechnologyuhh.github.io/GlassNICOLDataset/}{https://knowledgetechnologyuhh.github.io/GlassNICOLDataset/}} that was collected on the Neuro-Inspired COLlaborator (NICOL), a humanoid robot platform. The dataset consists of 7850 images recorded from five different cameras. We show that our trained baseline model outperforms state-of-the-art open-vocabulary approaches. In addition, we deploy our baseline model in an embodied agent approach to the NICOL platform, on which it achieves a success rate of 81\% in a human-robot bartending scenario.

\end{abstract}

\copyrightnotice

%%%%%%%%%%%%%%%%%%%%%%%%%%%%%%%%%%%%%%%%%%%%%%%%%%%%%%%%%%%%%%%%%%%%%%%%%%%%%%%%
\section{INTRODUCTION}
Transparent objects are everywhere, from households, healthcare assistance, and gastronomy to industrial and construction-site environments. Yet, these materials reflect only a very small fraction of visible light---most of it is scattered and passes through. From a computer vision perspective, the processing of these objects is challenging but crucial for a successful deployment in real-world robotics applications \cite{Sajjan2020cleargrasp}. As intelligent robots start to make their way into public spaces, where glass objects in the form of bottles and drinking glasses are integral items, future generations of service robots will be required to not only detect but also manipulate those materials. In addition, machine learning and data-driven methods require a large amount of training data to reach a suitable performance. Similar to robotic Sim2Real applications, existing research also proves the presence of a significant Sim2Real gap in the area of synthetic glass material generation \cite{Jiang2023, dai2022dreds, Sajjan2020cleargrasp}, while those approaches require sophisticated photorealistic rendering methods. %linked to high computational cost. 

Open‑vocabulary approaches are used for object detection in a plethora of embodied robotic setups, such as \cite{Ali2024, AAW25}. Embodied robotic agents, at a minimum, consist of a detection module, an action module, and an LLM utilized for high-level interaction and task planning. Recent approaches aim to fuse detection and language processing by utilizing vision-language models (VLMs) \cite{Elmira2024}. Being interactive by design, these robotic agents are typically situated in scenarios involving user interaction and often collaboration, leaving a potentially infinite corpus of object names to be detected. 
 
Well-known open-vocabulary object detection approaches such as Grounding DINO \cite{Liu2023GroundingDM}, YOLOWorld \cite{Cheng2024YOLOWorldRO}, and OWL-Vit \cite{Minderer2022owl} have revolutionized object detection by expanding the recognizable object classes without extensive re‑annotation. These methods are built upon end-to-end transformer‑based architectures. Global image features are extracted by decoder backbones and fused with language embeddings by utilizing deep metric learning. Despite their success, our experiments suggest that these models exhibit a significant decrease in performance when encountering transparent objects, commonly found in HRI scenarios. To address these challenges, we introduce a novel real-world dataset that was captured on the humanoid robot NICOL (Fig.~\ref{nicol}), including a new depth-based automated labeling method. The robot is equipped with a combination of RGB-D scanners and standard RGB cameras. Specific contributions of our work include:
\begin{itemize}
    \item A \textbf{novel real-world dataset} specifically designed for glass object detection, addressing the limitations of synthetic datasets and bridging the Sim2Real gap.
    \item An \textbf{automated labeling pipeline}, shown in Fig.~\ref{fig:pipe}, allowing a rapid annotation of data, leveraging existing visual foundation models.
    \item A \textbf{real-time integration} of the visual detector with a physical robot, showing its potential in HRI scenarios.
    \item A \textbf{humanoid bartender task} as a use case, where the robot accurately detects glassware and executes pouring motions, establishing a platform for controlled experiments in realistic, socially engaging contexts.
\end{itemize}

\section{RELATED WORK}

%In this section, we go over existing works from both computer vision and robotics areas, with a focus on existing datasets and their limitations.

\subsection{Transparent Material Perception}

%Robots need 3D depth information to perceive transparent objects. 
Missing or incorrect %depth 
RGB-D data %in the RGB-D captures 
for robotic perception
can be estimated %by leveraging 
using implicit functions and geometric priors~\cite{zhu2021implicit} and by balancing local and global depth features~\cite{li2023fdct}. The depth map can also be de-projected into a point cloud and processed with 
techniques like 3D CNNs to achieve depth completion~\cite{xu2021seeing}. 
Iterative refinement is also possible, with an attention module on transparent regions \cite{Zhai2024refine} or with indirect geometry representations~\cite{Tang2024RF}. 

RGB often contains information missing from depth data, which can be leveraged with transfer learning~\cite{Weng2020multimodal}. Cues can be extracted from the RGB to fill in depth observations using global optimization~\cite{Sajjan2020cleargrasp}. Incorporating affordance detection of the underlying object yields further improvement~\cite{Jiang2022aff}. Other approaches include multimodality, such as visual-tactile fusion~\cite{Murali2023Touch} and usage of polarization imaging~\cite{YU2024Polar}.

Recently, neural radiance fields (NeRFs), have emerged as a promising tool for transparent object perception. GraspNeRF introduces a multiview-based 6-DoF grasp detection system~\cite{Dai2022graspnerf}. Evo-NeRF extends this approach with real-time NeRF training and grasp adaptation~\cite{Kerr2022EvoNeRFEN}. Multiview methods enhance transparent object perception by integrating multiple viewpoints~\cite{wang2023mvtrans}. Domain randomization techniques have also been used to train models that generalize well to the real-world~\cite{Xompero2019MultiViewSE}. The common limitation of existing approaches is their unpredictable performance on unseen scenes, showing the need for the methodology of automated benchmark data aggregation and annotation without human labor \cite{Jiang2023}. 

\subsection{Existing Datasets}

The field of transparent object perception has long faced a scarcity of real-world datasets, with the majority of the work driven by synthetic data~\cite{Jiang2023, Tang2024RF, Lukei2024Track}. Synthetic datasets like Trans10k~\cite{Xie202010k}, SuperCaustics~\cite{Mousavi2021SuperCausticsRO}, and Dex-NeRF~\cite{IchnowskiAvigal2021DexNeRF} offer a vast number of images (ranging from 9k to 100k) but suffer from the Sim2Real gap~\cite{Josifovski2022sim2real}. Besides common causes like the lack of noise and imperfections presented in real-world environments, transparent objects provide a unique challenge due to their complex light interaction. In contrast, real-world datasets like TransCG~\cite{Fang2022transcg} and ClearGrasp~\cite{Sajjan2020cleargrasp} have attempted to address these limitations by collecting depth data using physical sensors. However, they lack diversity. Many datasets focus on isolated objects placed against controlled backgrounds, limiting their utility in highly cluttered environments~\cite{Jiang2023}. The process of manual ground-truth depth and segmentation annotation remains a significant bottleneck~\cite{Chen2022ClearPoseLT}. Few works, like the Toronto Transparent Object Depth dataset leverage automated annotation techniques~\cite{xu2021seeing}. %\HA{
Also, Liu et al.~\cite{liu2020} presented a multiview automated annotation approach. However, it is limited to isolated scenes and does not address the demanding challenge of dense transparent object detection and classification in realistic HRI scenarios, which our dataset is specifically designed for.
%}

Moreover, many existing large-scale RGB-D datasets~\cite{Lopes2022} do not specifically focus on transparent objects~\cite{Nguyen2024MCDDL}. We also note the existence of datasets like RGBP-Glass~\cite{Haiyang2022glass} and RGB-Thermal~\cite{huo2023therm}, introducing additional sensing modalities. Yet, these remain niche applications and do not fully address the generalization problem for robotic perception.

%\HA{Also, KeyPose~\cite{liu2020} presented a multiview annotation pipeline for transparent objects with accurate 3D labeling via RGB-D fusion and synthetic augmentations. However, their setup requires controlled environments and isolated scenes. In contrast, we target the more complex challenge of dense object detection and classification in realistic HRI scenarios.}
%does not directly address clutter or occlusion in realistic HRI scenarios, which our dataset is specifically designed for.

\begin{figure}
\vspace{5pt}
\centerline{\includegraphics[width=0.95\columnwidth]{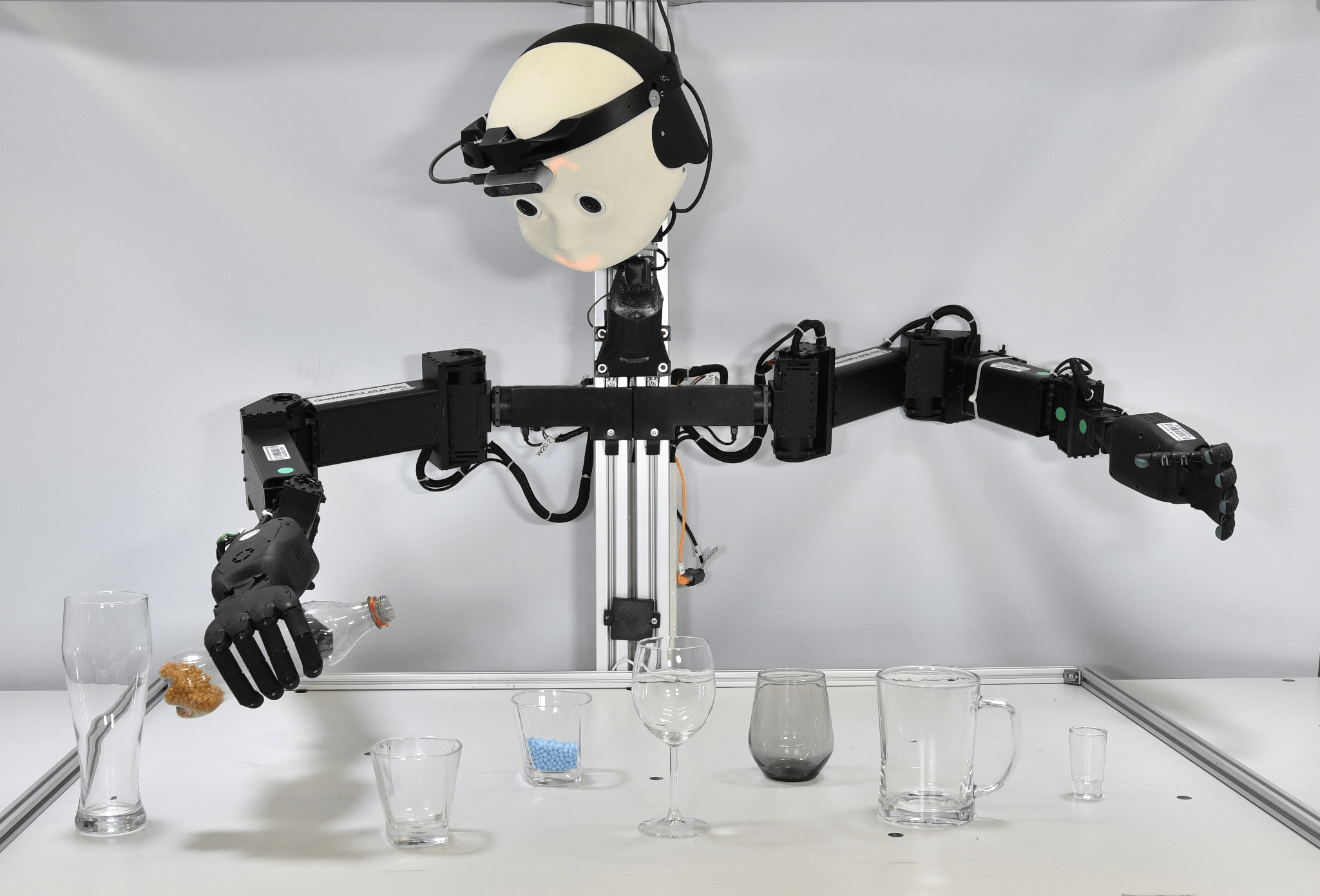}}
\caption{The NICOL humanoid robot used for down-stream task evaluation.}
\label{nicol}
\vspace{-4ex}
\end{figure}

\section{Methodology}

\begin{figure*}
    \vspace{5pt}
    \centering
        \begin{subfigure}[b]{0.24\textwidth}
            \centering
            %\includegraphics[width=0.95\textwidth]{imgs/depth.png}
            %\caption{Depth Map (Chalk Spray)}
            %\vspace{3mm}
        \end{subfigure}
        \begin{subfigure}[b]{0.24\textwidth}
            \centering
            \includegraphics[width=0.95\textwidth]{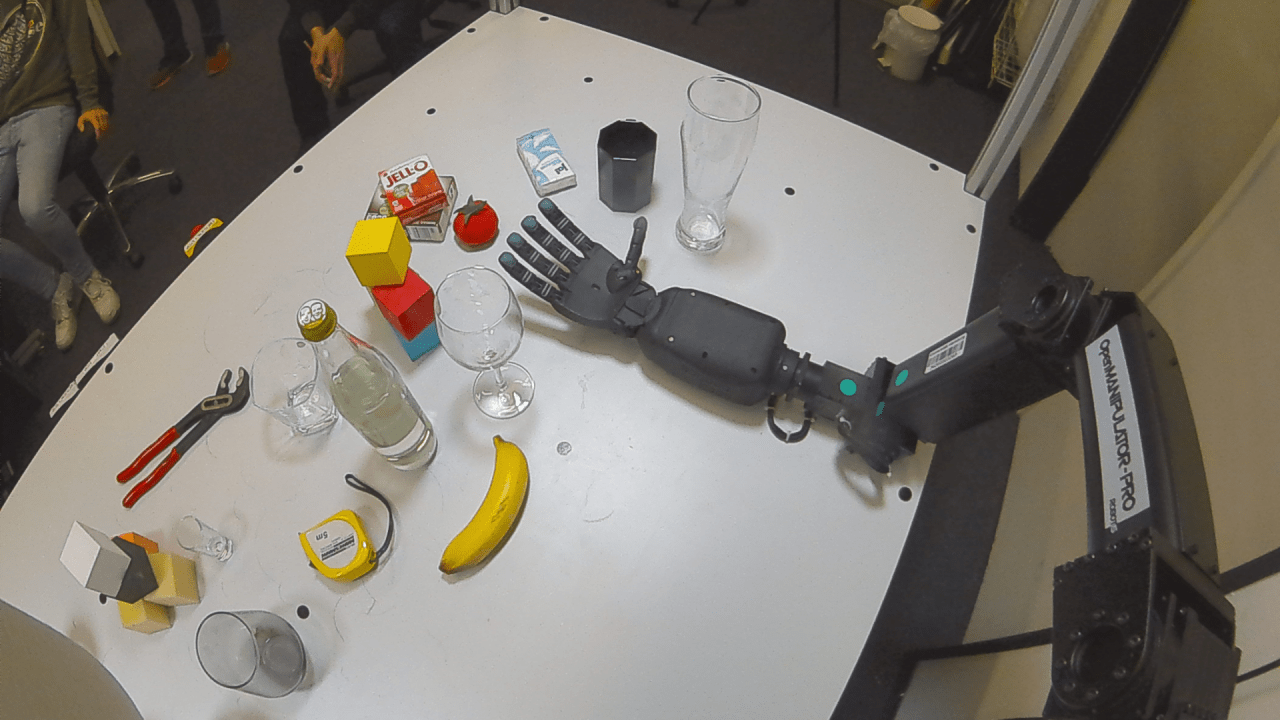}
            \caption{Eye Camera RGB Texture}
            %\vspace{3mm}
        \end{subfigure}
        \begin{subfigure}[b]{0.24\textwidth}
            \centering
            \includegraphics[width=0.95\columnwidth]{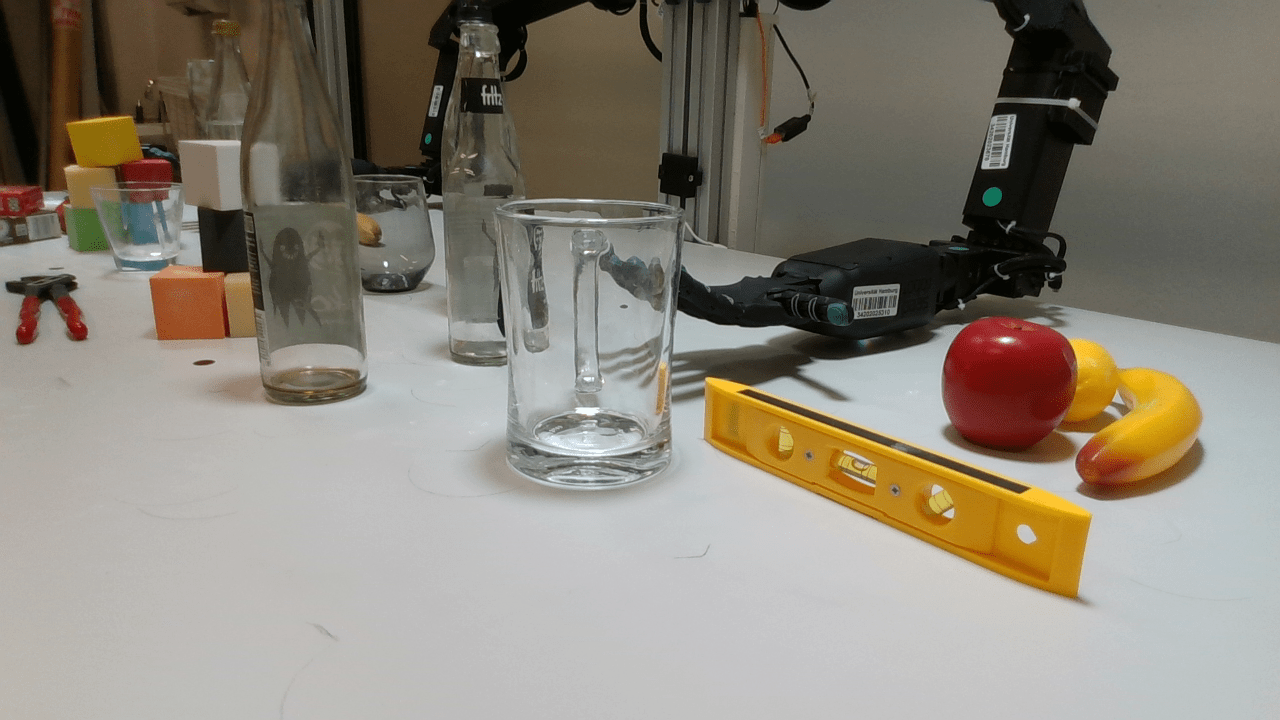}
            \caption{Side RealSense RGB}
            %\vspace{3mm}
        \end{subfigure}
        \begin{subfigure}[b]{0.24\textwidth}
        
            \centering
            
            \includegraphics[width=0.95\textwidth]{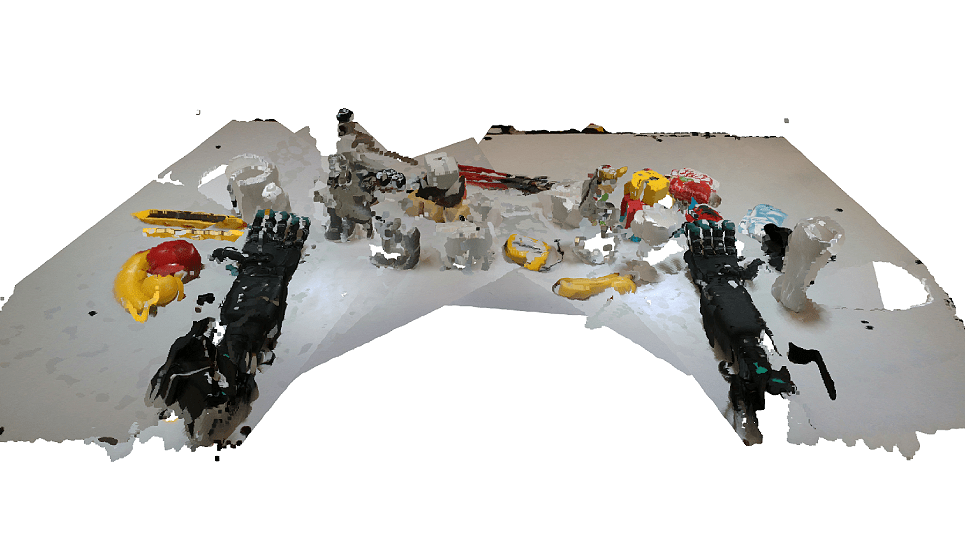}
            
            \caption{Fusion of all RGB-D Views}
            %\vspace{-9.5pt}
        \end{subfigure}
        %\begin{subfigure}[b]{0.32\textwidth}
        %    \centering
        %    \includegraphics[width=0.95\textwidth]{imgs/annotation.png}
        %    \caption{Auto-Generated Labels}
        %    \label{labels}
        %\end{subfigure}
        \begin{subfigure}[b]{0.24\textwidth}
            \centering
            \includegraphics[width=0.95\textwidth]{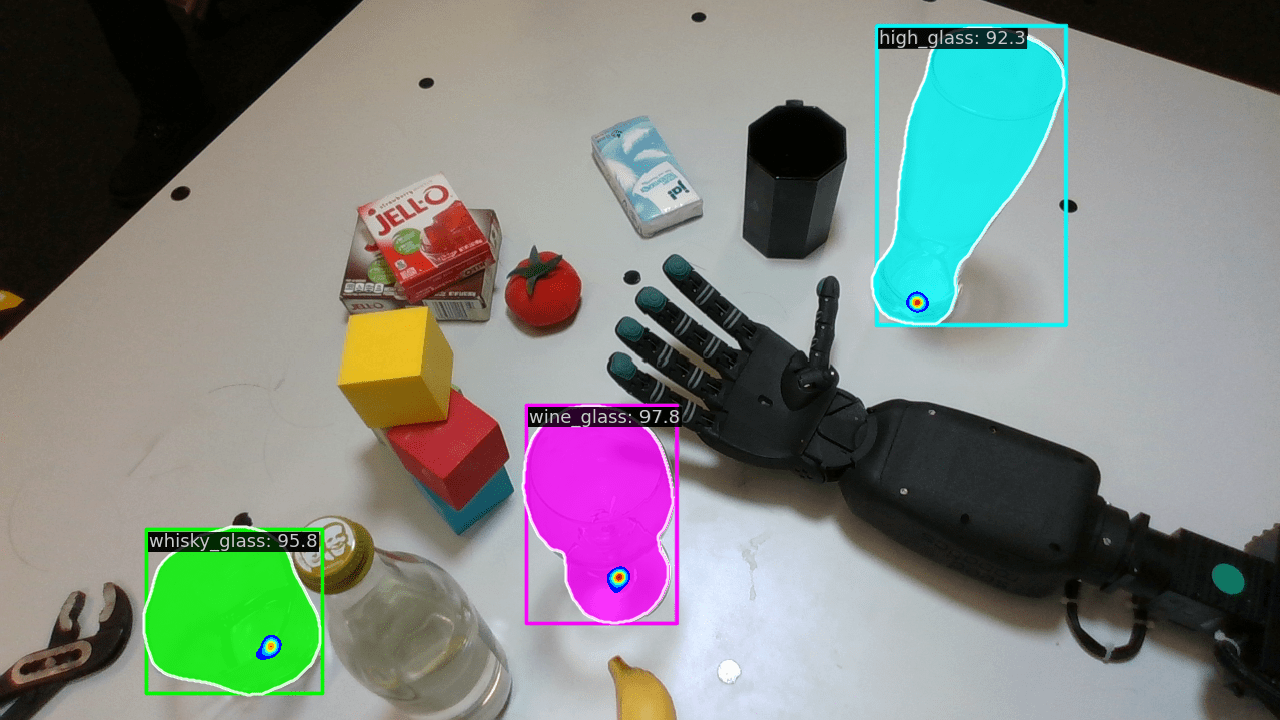}
            \caption{Network Prediction}% with Heatmap Bases}
            \label{heatmap}
            %\vspace{3mm}
        \end{subfigure}
        \caption{Different modalities present in our dataset, visualized on \texttt{scene\_145} from the validation split.}
    \label{textures}
    \vspace{-2ex}
\end{figure*}

\subsection{NICOL Platform and Camera Model}
%\todo{Jan}
\label{sec:camera_model}

The humanoid robot NICOL (see Fig.~\ref{nicol}) has two 8-DoF arms, each with an anthropomorphic tendon-driven 5-DoF hand. It has a 2-DoF 3D-printed head structure with an integrated facial expression interface. The platform is designed for manipulation and HRI scenarios and offers various sensors for different modalities. A two-meter-wide, one-meter-deep tabletop is attached in front of NICOL. The robot head contains two RGB fish-eye cameras with a 4k resolution. In our setup, a RealSense D435if camera is additionally mounted to NICOL's head, along with two RealSense cameras to the left and right front pillars of the robot's frame. 
Thus, our setup enables a complete depth perception of the objects on the table from multiple perspectives.

The movable head and large area on the table covered by NICOL's arms, combined with the fish-eye property of the RGB eyes, require a very precise camera calibration to allow for complex manipulation tasks such as pouring liquids. We use a semi-automated calibration method, which scans the table in front of NICOL from different angles to collect calibration data. Small calibration markers are attached to the tabletop that are automatically detected with a blob detection method. The detected positions in the camera image are manually mapped to the corresponding 3D ground truth positions by a human. The collected data is then used with the OpenCV camera calibration method \cite{2015opencv} to calculate the camera matrix $K$, the distortion coefficients $D$, as well as the rotation matrix $R$ and the translation vector $t$. The calibration has to be performed individually for each of the three cameras used in our setup. We reached a reprojection error of $\sim2~pixels$ for the RGB eye cameras and $\sim4.5~pixels$ for the RGB-D camera. The two static RealSense cameras were calibrated with the well-known chessboard calibration method. Our camera model utilizes the camera parameters to provide a ray-casting method between the table and the camera sensor. In other words, pixels from one camera can be cast to the corresponding 3D coordinate at the tabletop and, for example, back into one of the other cameras to mark a specific 3D coordinate in the images from all five cameras.

\subsection{Dataset Design}
\label{detadesign}
Recent approaches dealing with transparent object perception usually rely on synthetically rendered data \cite{Sajjan2020cleargrasp, dai2022dreds, wang2023mvtrans}. The amount of real-world data from scenes containing these objects is severely limited, in comparison to datasets with objects from opaque materials \cite{Lopes2022, Jiang2023}. This is especially true for scenes with cluttered environments and a mix of transparent and non-transparent objects. %\cite{Jiang2022aff}. 
We have designed a new benchmark dataset that is applicable to several computer vision and robotic tasks. Image frames are captured from the egocentric view of the NICOL robot \cite{10304130} performing various arm movements. Thus, complexity in various modalities is added as the dataset includes various degrees of occlusion, varying lighting conditions, and differing tabletop textures. 

Specifically, our dataset consists of scenes with a mix of objects placed on a table in front of a humanoid robot (see Fig. \ref{textures}). The collection process involves three separate passes:
\begin{enumerate}
    \item Capturing the scene with clean glasses.
    \item Placing 3D-printed green caps on top of the glasses for height measurement.
    \item Replacing glasses with identical instances sprayed by chalk spray, making them opaque for obtaining depth ground truth geometry.
\end{enumerate}

This design allows us to provide glass classification and detection labels, segmentation masks, and ground-truth depth measurements. Using the processing pipeline described in the following subsection, all annotations are created \textit{automatically, without human labor}. As mentioned earlier, our setup consists of three RealSense RGB-D cameras, one attached to the robot's head and two at the sides of the table. Also, there are two fisheye RGB cameras, one in each eye of the robot. Using this setup, we captured 100 different scenes, with variations in table color and texture, lightning conditions, robot movements, and scene compositions. For each scene, there are 25 different views from the robot, acquired by a rotation in the neck joint. All cameras were calibrated and registered, allowing a projection of detection labels from the head-mounted RGB-D sensors to all other views.

The sum of frames acquired from all five cameras (2 static RGB-D cameras and 25 images times 3 head cameras) results in a number of 77 frames per scene and 7700 frames in the training dataset in total. There are six different glass types present, each with its own class, namely: \textit{high beer glass, beer glass with a handle, wine glass, water glass, whiskey glass,} and \textit{shot glass}. For each type, we also provide a high-quality 3D model usable in surface reconstruction tasks~\cite{Li2020ThroughTL}. Together with ground truth depth observations from the three RGB-D scanners, this can also be used in monocular depth estimation~\cite{Yang2024} and depth completion tasks~\cite{Fang2022transcg}. Our main use case of the data in this paper is the detection and classification of a glass type. In a more general perception task, different glass types can be omitted, and all labels can be treated as general \textit{drinking glass} annotations. 

Our dataset intentionally contains various unlabeled bottles, serving as negative examples during training. These objects are made of transparent material but are not typically classified by humans as drinking glasses. In addition to the 7700 frames used for training and validation, we captured separate testing data containing 150 frames: 75 frames captured using the same approach as the training data and another 75 frames captured during the bartender experiment described in Sec.~\ref{bartendertask}, thus having a completely different character compared to the training data. The labels of the test data were created manually to mitigate the effect of label noise introduced by our automated pipeline described below.

\subsection{Auto-Labeling Pipeline}

The labeling process uses depth sensing, color verification, and object detection to create accurate segmentation masks and bounding boxes. Initially, depth images are converted into a 3D point cloud%representation
, denoted as \(P = \{ p_i \in \mathbb{R}^3 \} \) where each point \( p_i = (x_i, y_i, z_i) \) represents spatial coordinates obtained from the RealSense camera mounted on the robot's head. The primary surface in the scene, corresponding to the table, is estimated using RANSAC-based plane fitting, yielding the plane parameters \( a, b, c, d \). Points deviating significantly from this plane are identified as potential objects and subsequently clustered using a density-based clustering algorithm from Open3D~\cite{Zhou2018Open3DAM}, ensuring each cluster represents a discrete candidate object. Next, the height of the clusters is computed relative to the table plane using the formula: 

\begin{equation}
    h = \frac{|d' - d|}{\sqrt{a^2 + b^2 + c^2}},
\end{equation}

%\[
%h = \frac{|d' - d|}{\sqrt{a^2 + b^2 + c^2}},
%\] 
where \( d' \) is the cluster's plane offset.
%\note{
Clusters with large height deviations from known glasses are discarded. The remaining candidates are given classes based on the closest known glass height, then 
%} 
%We discard clusters with large height deviations from known glasses. The remaining candidates are 
projected onto the 2D image plane using intrinsic camera parameters, transforming 3D coordinates into pixel locations \( (u, v) \) via perspective projection:
%\begin{equation}
%    \begin{bmatrix} u \\ v \\ 1 \end{bmatrix} \sim K %\begin{bmatrix} x \\ y \\ z \\ 1 \end{bmatrix}
%\end{equation}
\begin{equation}
    \begin{bmatrix} u & v & 1 \end{bmatrix}^{T} \sim K \begin{bmatrix} x & y & z & 1 \end{bmatrix}^{T},
\end{equation}

%\[
%\begin{bmatrix} u \\ v \\ 1 \end{bmatrix} \sim K \begin{bmatrix} x \\ y \\ z \\ 1 %\end{bmatrix},
%\]
where \( K \) is the intrinsic calibration matrix of the camera. To further filter %relevant 
instances, color verification is performed in the CIELAB color space, ensuring that the detected object caps match the expected green hue. A final verification step uses YOLO-World, a deep-learning-based object detector, to confirm the detected object has glass-like features, mitigating false positives from other green objects in the scene~\cite{Cheng2024YOLOWorldRO}. As each of these steps works with different data modalities and false-positive error profiles, we achieve a variant of a cascaded filtering approach by fusion of outputs from several algorithms
~\cite{klein1999Fusion}.
After obtaining the final candidates, points from the blobs are used as samples for the Segment Anything Model (SAM)~\cite{Ravi2024}.
%, generating object masks for precise annotations. 
%\note{
As shown by our final data, SAM achieves reliable segmentation performance in cluttered environments when guided by our precise sample points. To improve accuracy, we discard overly wide masks, often caused by merged neighboring objects.
%}
%The segmentation masks \( M_i \) are converted into polygons, defining the object's shape in the image space. 
To finalize the object instance label, we derive bounding boxes \(B_i = (x, y, w, h) \) from the masks.
%We derive bounding boxes \(B_i = (x, y, w, h) \) from these masks, finalizing the object instance label. 
An overview of our labeling pipeline and an example annotation were presented earlier (see Fig.~\ref{fig:pipe}).

\subsection{Baseline Detector}

To demonstrate the advantages of our novel dataset, we use it to train an object detector and perform use case scenarios and experiments. For real-time detection performance, we opted for an RTMDet architecture in its medium-size variant~\cite{Lyu2022}. Naturally, different one-stage architectures, like the YOLO family~\cite{sapkota2025}, could be employed for this task.

The network variant chosen incorporates various training strategies. Specifically, the network employs a CSPNeXt backbone with a P5 architecture, a deepen factor of 0.67, and a widen factor of 0.75, enhanced with channel attention and synchronized batch normalization. The neck is a CSPNeXtPAFPN structure with two CSP blocks~\cite{Wang2019CSPNetAN}, facilitating multi-scale feature aggregation~\cite{Lyu2022}. As proposed by the original authors, the head of the network calculates three loss functions. A Quality Focal Loss (\(\beta = 1.0\), loss weight = 1.0) for classification~\cite{Li2021}, Complete IoU (CIoU) Loss (loss weight = 2.0) for bounding box regression~\cite{Zheng2022}, and Dice Loss (loss weight = 3.0, 
%\(\epsilon = 5 \times 10^{-6}\), 
reduction = ``mean'') for mask prediction~\cite{Lyu2022}. To increase robustness and grant better generalization for different scenarios, heavy data augmentation is employed. This includes techniques such as Cached Mosaic~\cite{Bochkovskiy2020YOLOv4OS}, Cached MixUp~\cite{Zhang2017mixupBE}, Random Resize, Random Crop, and YOLOX-style HSV augmentation~\cite{Ge2021YOLOXEY}. 

The training is performed from a pre-trained checkpoint~\cite{Lyu2022}, which we fine-tune on our dataset. The learning rate is scheduled using a combination of Linear Warmup (start factor = 0.001, epochs 0–50) and MultiStepLR with decay at epochs 100, 200, and 400. The optimizer is AdamW~\cite{loshchilov2018decoupled} with a learning rate of 0.01 and a weight decay of 0.05. The model is trained for 500 epochs with a batch size of 8, and evaluated every 50 epochs using the COCO metric for both bounding box and segmentation performance on 5 validation scenes (a 5\% split from our 100 scenes). 

\subsection{Glass Base Points}
\label{sec:glass_base_points}
%Without modifications, the RTMDet-m identifies transparent objects in 2D  by predicting bounding boxes and segmentation masks. 
To improve the robotic arm guidance without heavy modification of the detector architecture or lifting to 3D \cite{Wang20233D}, we calculate 2D glass base positions separately. Given these locations and the known heights of the glasses from the classification, it enables robot pouring without exact 3D positions. We provide annotations for them in the dataset. By calculating the average position 
of each 3D-printed cap, we obtain point \( \mathbf{p} = (x, y, z)^\top \) for each glass and collect them into a set \( \mathbf{C} \). Taking the normal vector of the table \( \mathbf{n} = (a, b, c)^\top \) and its normalized version \( \mathbf{\hat{n}} \), the perpendicular distance to the table for each \( \mathbf{p} \in \mathbf{C} \) is given by:

\begin{equation}
    d_{\text{proj}} = \frac{ax + by + cz + d}{\|\mathbf{n}\|}.
\end{equation}

The projection of the point onto the table is computed as  \(
\mathbf{p}_{\text{proj}} = \mathbf{p} - d_{\text{proj}} \mathbf{\hat{n}},
\). To obtain 2D image coordinates, we use perspective projection with the camera matrix \( K \). Taking 2D coordinates as centers of small bounding boxes with a fixed size, we introduce them as a separate \texttt{keypoint} class into the COCO object detection format.

Finally, we propose a modification of the detector head to obtain heat-map of glass base points. Before applying Non-Maximum Suppression \cite{Stauffer1999NMS}, we extract all proposals of the \texttt{keypoint} class. We calculate the center location \( (x_c, y_c) \) for each proposal and place Gaussian kernel \( G(x, y) \) of size \( k \times k \) (in our experiments \( k = 15 \)) at these locations:

\begin{equation}
    G(x, y) = \frac{1}{2\pi\sigma^2} \exp \left( -\frac{(x - x_c)^2 + (y - y_c)^2}{2\sigma^2} \right).
\end{equation}

The contribution of each kernel is weighted by the confidence score \( s_i \) of the corresponding bounding box. 
The final heatmap \( H(x, y) \) is obtained by summing all contributions:
\begin{equation}
    H(x, y) = \sum_{i} s_i G_i(x, y) .
\end{equation}

2D base points are extracted by identifying local maxima in the heatmap within each detected glass bounding box (see Fig.~\ref{heatmap} for an example with heatmap overlaid on detected glasses). 2D points can also be calculated with the approach in Section \ref{sec:pouring_motion}.
Full 3D localization using the depth maps provided in that dataset is proposed for future work.

\begin{promptfloat}
\vspace{5pt}
\begin{prompt}{bartender_prompt}{A summarized version of our system LLM prompt of the bartender task with an example of the pouring wine and pouring beer functionalities.}
You are deployed to function as a bartender. There are glass objects on the table. You should always respond as a bartender and offer available services like a bartender. When serving drinks, follow these rules:\\[1ex]
    1. Pouring Wine:\\
		- Identify the correct bottle.\\
		- Identify the correct glass.\\
		- Example: $<$pour\_wine(wine glass)$>$.\\
		If there is only one wine glass, pour immediately into it without asking the user. If there are multiple wine glasses, ask the user which they prefer before pouring. \\
    2. Pouring Beer:\\
		- Identify the correct bottle.\\
		- Identify the correct glass.\\
		- Example: $<$pour\_beer(beer glass)$>$.\\
        - Example: $<$pour\_beer(high beer glass)$>$.\\
		If there is only one beer glass, pour immediately into it without asking the user. If there are multiple beer glasses, ask the user which they prefer before pouring. \\
        %... \\[1ex]
        You can fill a glass only once. When you pour into a glass, you should remember which glass is already filled. Always prioritize serving drinks accurately and safely. Respond to the user like a professional bartender.

\end{prompt}
\vspace{-5ex}
\end{promptfloat}

\subsection{LLM Integration: Bartender Task}
\label{sec:llm_integration}

The NICOL platform supports an LLM integration for embodied agents in real-world robotic tasks~\cite{allgeuer2024_chattyrobots}. We utilize the grounded LLM to implement a robot bartender scenario, leveraging the LLM's reasoning to connect the robot's sensory \textit{perception} (identifying glass objects) and robot's physical \textit{actions} (precise arm manipulation for pouring). Our task facilitates natural interaction with users since the bartender scenario inherently involves social dynamics like recognizing drinkware, interpreting user requests, and responding appropriately. %\HA{
Moreover, bartending tasks have proven to be engaging testbeds for social HRI, as shown in systems like BRILLO~\cite{rossi2025}.
%} 
Although our implementation focuses on verbal communication, the task can be extended to incorporate context-driven and multimodal cues, including nonverbal communication. The LLM prompt is in Prompt~\ref{prompt:bartender_prompt}.

As shown in our previous work~\cite{Ali2024}, the concept of procedural memory in LLMs is effective for rapid acquisition of text-based skills in robots, enabling them to perform adaptive and context-aware reasoning. We apply this concept to implement interactive use cases in the bartender task, where NICOL assists users by serving drinks. While not exhaustive, our scenario highlights the potential of integrating our proposed glass dataset and detector into a robotic application. The scenario consists of the following use cases:

\begin{enumerate}
    \item Action-Object Alignment: The robot accurately maps the beverage pouring action to appropriate glassware in a way that aligns with user expectations and common conventions (e.g., pouring wine into a wine glass).
    \item Resolving Ambiguities: The robot identifies potential ambiguities in object selection and seeks clarification through user corrective feedback. For example, if the user orders beer and multiple beer cups are available, the robot intuitively prompts the user to specify their preferred glass type, while providing a list of options.
    
    \item Context-Driven Action Sequence: The robot uses LLM context memory to maintain continuity across the interaction, ensuring sequential tasks are carried out in logical order. For example, if the user orders a beer and then sequentially asks for another, then the robot can reason about which glass it already filled and choose the appropriate glass for the next pouring action. %accordingly.
\end{enumerate}

\subsection{Robot Integration: Beverage Pouring Action}
\label{sec:pouring_motion}
The B\'ezier curve-based motion planner described in our previous work \cite{10802010} is utilized to design a pouring motion for the robot bartender task. The action picks up an already open bottle from a fixed position, moves it towards the opening of the glass, pours particles into it, and returns the bottle to its origin. Careful and precise pouring motions are required to prevent spilling particles and to avoid damaging the glasses\footnote{No glass was harmed in any of our experiments.}. The RealSense camera attached to the head of NICOL is used for this application, but it is also possible to use the left or right eye cameras. We use 3D-printed particles to imitate liquid, as it prevents damage to the electrical hardware. 

We use the projection approach of the camera model described in Sec.~\ref{sec:camera_model} to determine the glass coordinates in 3D space. First, the center of the bounding box bottom edge~$b$ is projected from the image plane onto the tabletop. Thus, the approximately nearest point between the camera origin and the glass bottom is calculated. Since the given coordinate is located at the outer hull of the glass, an offset $o_{i}$ is added that takes the height and diameter of the i-th glass class into account. The center position of the glass opening changes its relation to the detected bounding box bottom center with respect to the glass's position in the table plane and the height of the glass. The beverage pouring motion has to account for those spatial changes through a second dynamic pouring offset $p_x$ and $p_y$ that scales linearly with the specific x- and y-axis positions, as shown in Eq.~\ref{eq:x_offset_pouring}:
\begin{equation}
\label{eq:x_offset_pouring}
\begin{split}
    p_x = \epsilon \cdot p_x^{min} + (\epsilon \cdot \tau) p_x^{max}\\
    p_y = \gamma \cdot p_y^{min} + (\gamma \cdot \tau) p_y^{max} .
\end{split}
\end{equation}

In the prior equation, suitable offsets for the smallest glass class $p_x^{min}$ and $p_y^{min}$ are adjusted with the linear scaling factor for the x-axis position $\epsilon$ and the y-axis position $\gamma$. In addition, suitable offsets for the highest glass class $p_x^{max}$ and $p_y^{max}$ are added by adjusting them with the linear scaling factor for the glass height $\tau$. As the height offset has shown to be influenced by the x- and y-position of the glass in our preliminary experiments, it is multiplied by the scaling factor for the corresponding axis $\epsilon$ or $\gamma$. The pouring offsets for the smallest and highest glass classes $p_x^{min}$, $p_y^{min}$, $p_x^{max}$, and $p_y^{max}$ are determined by placing the smallest and highest glass at the maximum x- and y-position of the pouring workspace and manually tuning the offsets. The primitive scales well to intermediate glass classes. The normalized scaling factors $\epsilon$, $\gamma$, and $\tau$ are calculated as in Eq.~\ref{eq:scaling_factors}:
\begin{equation}
\label{eq:scaling_factors}
\begin{split}
    \epsilon =& \frac{x_n - x_{min}}{x_{max} - x_{min}},\hspace{5pt} \gamma = \frac{|y_n| - y_{min}}{y_{max} - y_{min}}\\
    &\hspace{20pt}\tau = \frac{h_n - h_{min}}{h_{max} - h_{min}},\\
\end{split}
\end{equation}

where the constants $x_{min}$, $x_{max}$, $y_{min}$, and $y_{max}$ are the workspace bounds. Similarly, $h_{min}$ and $h_{max}$ are the height of the smallest and highest glass class.

\section{Experiments and Evaluation}
We conduct two experiments to evaluate our system. First, we evaluate the performance of our fine-tuned RTMDet glass classifier using our proposed dataset and compare it against state-of-the-art off-the-shelf open-vocabulary object detectors. %\note{
Additionally, we verify our claim regarding the insufficiency of synthetic data caused by the challenging glass material. Using 3D models of glasses provided within our dataset, we train the network on data rendered by the SuperCaustics framework \cite{Mousavi2021SuperCausticsRO} and include its performance in the evaluation.
%} 
In the second experiment, we integrate our glass classifier with an LLM-powered embodied agent using NICOL and perform an end-to-end system evaluation in a bartender-like Human-Robot scenario. Next, we show the evaluation metrics and results of each experiment.

\subsection{Transparent Object Detection}

We evaluate the detector on the test data using the two standard \textit{COCO} evaluation metrics \textit{Average Precision} and \textit{Average Recall} at IoU from $0.5$ to $0.95$. Using data from the eye cameras in addition to the captures from the head-mounted RealSense during training slightly increases the performance. The values are calculated for two cases. In the first case (highlighted in Table~\ref{detectotab}), we use the class names described in Sec.~\ref{detadesign} as prompts to the open-vocabulary detectors. The shown values of AP and AR are the mean across all glass types (sometimes referred to as \textit{mAP} and \textit{mAR}, respectively). %Obviously, this is not completely fair, as 
It should be mentioned that existing detectors are unaware of our classification of glass types. There is an ambiguity even for humans, such as the difference between a \textit{whiskey} and a \textit{water} glass. Therefore, in the second case, we treat all instances as a general \textit{drink glass} class. In preliminary experiments, we have also tried other prompts, such as \textit{transparent drink container}, \textit{drinking glass}, or \textit{drinkware}, with worse results. The testing split, as the rest of our data, does not contain any other objects classifiable as a \textit{drink glass} by a human. We have also experimented with image-guided prompting with a template glass image. In such a configuration, the detectors preferred to look for other containers in a similar pose (such as from a top-right view), rather than focusing on the material and type. This caused an even higher false positive rate, with detections of objects like \textit{ceramic cups} in a pose similar to the template.

The higher AP of our pretrained model, even in the general \textit{drink glass} setting, supports our hypothesis that existing open-vocabulary detectors do not grasp the concept of a drink container made from glass. In their latent space, it seems confused with containers made from other materials like \textit{metal cans}. They also cannot distinguish between transparent containers like bottles and drinking glasses, which clearly serve different functions. Taking a further step towards glass type classification makes existing detectors unusable for our scenario. In terms of AR, existing models perform more appropriately. They do not have many false negatives, generally selecting everything that looks like a rounded container. 

\begin{table}[]
\vspace{1ex}
\centering
\caption{Comparison of glass detection using zero-shot detectors and transfer-learned lightweight model}
\begin{tabular}{lcccc}
\toprule
\textbf{Method} & \multicolumn{2}{c}{\textbf{AP@[0.5:0.95]}} & \multicolumn{2}{c}{\textbf{AR@[0.5:0.95]}} \\  & 
\begin{tabular}[c]{@{}c@{}}general\\ class\end{tabular} & 
\begin{tabular}[c]{@{}c@{}}glass\\ types\end{tabular} & 
\begin{tabular}[c]{@{}c@{}}general\\ class\end{tabular} & 
\begin{tabular}[c]{@{}c@{}}glass\\ types\end{tabular} \\ \midrule
OWL-Vit\cite{Minderer2022owl}                      & 0.403 & \cellcolor{pink!25}0.014          & 0.671  & \cellcolor{pink!25}0.112          \\
G-DINO\cite{Liu2023GroundingDM}                    & 0.638 & \cellcolor{pink!25}0.102          & 0.829  & \cellcolor{pink!25}0.357          \\
YOLO-World\cite{Cheng2024YOLOWorldRO}              & 0.707 & \cellcolor{pink!25}0.191          & 0.844  & \cellcolor{pink!25}0.413          \\
\begin{tabular}[c]{@{}c@{}}*RTMDet-M\cite{Lyu2022}
\\ (synth data \cite{Mousavi2021SuperCausticsRO})\end{tabular}                    & 0.224 & \cellcolor{pink!25}0.145 & 0.475  & \cellcolor{pink!25}0.290  \\ 
\begin{tabular}[c]{@{}c@{}}*RTMDet-M\cite{Lyu2022}
\\ (head RGB only)\end{tabular}                    & 0.774 & \cellcolor{pink!25}0.713 & 0.850  & \cellcolor{pink!25}0.801  \\ 
\begin{tabular}[c]{@{}c@{}}\textbf{*RTMDet-M}\cite{Lyu2022}
\\ \textbf{(all RGB views)} \end{tabular}                    & \textbf{0.786} & \cellcolor{pink!25}\textbf{0.718} & \textbf{0.854}  & \cellcolor{pink!25}\textbf{0.809}          \\ \bottomrule
\multicolumn{4}{l}{\scriptsize * Our fine-tuned models.} \\
\end{tabular}
\label{detectotab}
\vspace{-4ex}
\end{table}

%\begin{figure}
%\centerline{\includegraphics[width=0.9\columnwidth]{imgs/nicol.JPG}}
%\caption{NICOL platform used for down-stream task evaluation.}
%\label{nicol}
%\end{figure}

\subsection{Robot Integration: Bartender Task}
\label{bartendertask}
In this experiment, we use the LLM integration (see Sec.~\ref{sec:llm_integration}) and beverage pouring motion (see Sec.~\ref{sec:pouring_motion}) to evaluate the performance of our robot bartender task, deployed on the physical robot hardware. We evaluate our system in an end-to-end manner, i.e., a user (an experimenter) interacts with the system in a way that resembles a real-world scenario. Our experimental setup consists of NICOL with six glass objects placed on the table -- a \textit{shot glass}, \textit{wine glass}, \textit{whiskey glass}, \textit{water glass}, \textit{beer glass}, and \textit{high beer glass}. Additionally, two plastic bottles -- a \textit{wine bottle} and \textit{beer bottle} -- are placed on each side of the table and filled with artificial liquid made from 3D-printed round-shaped particles with distinct colors. The pouring workspace has a width of 55 cm and a depth of 35 cm. The positions of the bottles were swapped halfway through the experiment for a fair comparison of the pouring by each arm. The user stands across the table and issues verbal commands through an external microphone, processed via a Whisper integration. 

Each interaction consists of the following: the user orders wine, then a beer, followed by another beer request. We select these interactions to assess the system's ability to adapt to different beverage types, resolve ambiguities in glassware selection, and ensure a logical action sequence. This follows as a three-phase process: 1)~ User Order Recognition: The system processes the user's order and determines the appropriate drink and glassware, through action-object mapping or user preference. For example: if the user says \textit{``I'd like a glass of wine''}, the robot must infer that wine should be poured into a wine glass, 2)~Reasoning \& Decision Making: Based on user input, the robot reasons about the correct bottle and glass object. Any ambiguity is clarified through user feedback. For example: if the user says \textit{``Can I get a beer for my friend?''}, the robot must decide the correct glassware based on context and prior interactions (which glass is already filled), 3)~Action Execution: The robot performs a pouring action correctly to serve the request. We use OpenAI GPT‑4o mini and reset the chat after each interaction. 

\begin{figure}
\vspace{5pt}
\includegraphics[width=1\linewidth]{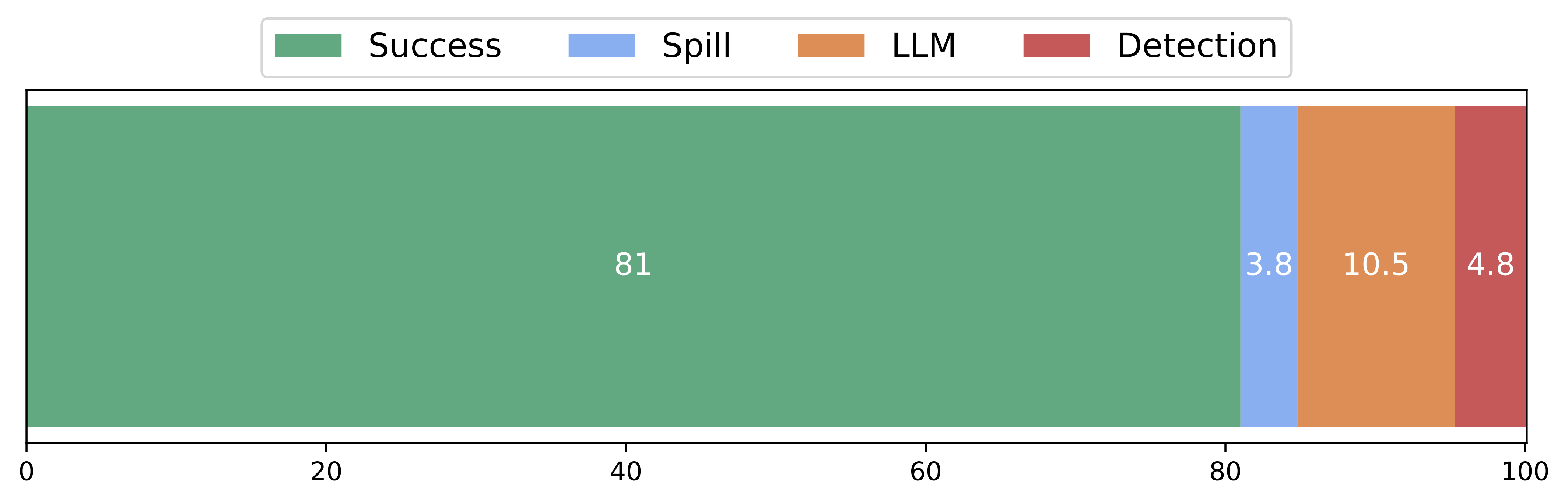}
\caption{Results of our end-to-end evaluation of the NICOL robot bartender task, highlighting success rates, spilling occurrences, errors in LLM reasoning, and glass detection.}
\label{pouring_experiment}
\vspace{-3ex}
\end{figure}

We conducted 35 trials, meaning that 105 pouring actions were executed in total. We use the \textit{Success} rate of the end-to-end pouring action as a metric to evaluate our experiment. However, we report the average error rate in the following categories: \textit{Spill} referring to slight spilling of particles outside the intended glass, \textit{LLM} representing LLM reasoning inaccuracies, and \textit{Detection}, i.e., errors in the glass detector (cf. Fig.~\ref{pouring_experiment}). Each pouring motion is considered successful if the robot correctly pours the target drink into the target glass without any particles spilled. Our system achieves a success rate of 81\% in the pouring action across the two robot arms, while only 3.8\% of the cases exhibited minor spills of particles outside the designated glass, demonstrating the suitability of our approach for precise pouring in a bartender task. Our LLM integration showed high resilience in generating correct robot actions across the different use cases: action-object mapping, ambiguity resolution, and logical action sequencing. In 10.5\% of the cases, user correction was needed due to reasoning inconsistencies, such as suggesting the wrong glass for a specific drink. However, only one instance was recorded, where the LLM generated an unintended action, mistakenly handing over the object instead of pouring into it, leading to a failure case. Only 4.8\% of the cases resulted in failures due to occasional glass detector errors, such as failing to detect the beer glass.

\section{Discussion}
As can be seen in Table~\ref{detectotab}, our new benchmark dataset helps to push the current state-of-the-art detector boundaries. Zero-shot models like YOLO-World\cite{Cheng2024YOLOWorldRO}, OWL-Vit \cite{Minderer2022owl}, and GroundingDINO\cite{Liu2023GroundingDM} are widely used for opaque models, but their performance deteriorates on concepts of transparency, glass material, and drink-purposed containers. On the other hand, our processing pipeline allows us to train a lightweight model from the RTMDet \cite{Lyu2022} family, achieving a high precision with a smaller rate of false positives. This proves the fact that reliable recognition of glass objects is far from achieved with general models. The glass detection system has shown high performance when integrated with the physical robot, achieving an overall success rate of 81\% in a bartending scenario. While the results show that there is still room for improvement in our object detection and robotic action modules, the biggest error source originates from the LLM utilized for high-level planning, making more than 50\% of the failure trials. Unfortunately, the glass base keypoints described in Sec.~\ref{sec:glass_base_points} oscillated too much to be applicable in the bartending agent in preliminary experiments. However, 3D planning approaches seem to be a promising area for future research. Our system already shows high performance by planning the motion based on the 2D image perception of the scene. A 3D perception approach, e.g., by utilizing the chalk samples delivered with our dataset for depth estimation, would thus still be capable of improving it.

\section{CONCLUSION}
This paper fills a critical gap in glass object perception by providing a novel real-world dataset. The proposed auto-labeling method reduces the human effort for data aggregation in similar tasks to an absolute minimum. We contribute three distinct open-source repositories: the glass dataset, the glass detector model and experiment code, and the NICOL camera model. The quality of our dataset and camera model allows for very precise projections between different camera perspectives. Our glass detector outperforms SOTA open-vocabulary approaches and reaches a baseline success rate of 81\% when integrated with our embodied bartending agent. Future work will include expanding the dataset with more diverse scenes and improving robotic manipulation strategies through enhanced LLM integration. The most interesting next step for the action module is a proprioceptive 3D planning approach that can dynamically adapt to spatial changes, like a user moving a glass while pouring, fostering seamless Human-Robot collaboration in real-world tasks.

\addtolength{\textheight}{-2cm}
\section*{Acknowledgements}
We thank Svorad Štolc from Photoneo for external consultations of the vision part, Philipp Allgeuer for his contribution to the NICOL robot, and Connor Gäde for his contribution to the bartending experiment.

%\addtolength{\textheight}{-0.4cm}   % This command serves to balance the column lengths

%%%%%%%%%%%%%%%%%%%%%%%%%%%%%%%%%%%%%%%%%%%%%%%%%%%%%%%%%%%%%%%%%%%%%%%%%%%%%%%%

%%%%%%%%%%%%%%%%%%%%%%%%%%%%%%%%%%%%%%%%%%%%%%%%%%%%%%%%%%%%%%%%%%%%%%%%%%%%%%%%

%%%%%%%%%%%%%%%%%%%%%%%%%%%%%%%%%%%%%%%%%%%%%%%%%%%%%%%%%%%%%%%%%%%%%%%%%%%%%%%%
%\section*{APPENDIX}


%\section*{ACKNOWLEDGMENT}

%The work presented in this paper was carried out in the framework of the TERAIS project, a Horizon-Widera-2021 program of the European Union under the Grant agreement number 101079338. The results were obtained using the computational resources procured in the project National competence centre for high performance computing (project code:~311070AKF2) funded by European Regional Development Fund, EU Structural Funds Informatization of society, Operational Program Integrated Infrastructure.

%%%%%%%%%%%%%%%%%%%%%%%%%%%%%%%%%%%%%%%%%%%%%%%%%%%%%%%%%%%%%%%%%%%%%%%%%%%%%%%%
\def\url#1{}
\bibliographystyle{IEEEtran}
\bibliography{references}

\end{document}